\pdfoutput=1
\documentclass{article}
\usepackage{spconf,amsmath,epsfig}
\usepackage{amssymb}
\usepackage{amsfonts}
\usepackage{float}
\begin{document}\sloppy

% Example definitions.
% --------------------
\def\x{{\mathbf x}}
\def\L{{\cal L}}

% Title.
% ------
\title{Decoupling Semantic Context and Color Correlation
with multi-class cross branch regularization}
%
% Single address.
% ---------------
\name{Vishal Keshav, Tej Pratap GVSL}
\address{Samsung Research Institute, Bengaluru}

\maketitle

\begin{abstract}
This paper presents a novel design methodology for architecting a light-weight and faster DNN architecture for vision applications. The effectiveness of the architecture is demonstrated on Color-Constancy use case an inherent block in camera and imaging pipelines. Specifically, we present a multi-branch architecture that disassembles the contextual features and color properties from an image, and later combines them to predict a global property (e.g. Global Illumination). We also propose an implicit regularization technique by designing ’cross-branch regularization’ block that enables the network to retain high generalization accuracy. With a conservative use of best computational operators, the proposed architecture achieves state-of-the-art accuracy with ‘30X’ lesser model parameters and ‘70X’ faster inference time for color constancy. It is also shown that the proposed architecture is generic and achieves similar efficiency in other vision applications such as Low-Light photography.
\end{abstract}
\begin{keywords}
Illumination estimation, Low Light photography, Multi-branch architecture, Regularization with soft-parameter sharing
\end{keywords}
\section{Introduction}
\label{sec:intro}

In the contemporary world, it has become ubiquitous to realize most of the vision-based tasks with Deep Neural Networks (DNNs) to achieve higher accuracy. This success led to the wide applicability of DNNs for the camera, image, and video applications like color constancy\cite{forsyth1990novel}, image de-noising\cite{burger2012image}, low light enhancement\cite{mayer1997learning} and image de-hazing\cite{cai2016dehazenet}. Since color is an important cue in many such vision applications, it is essential to provide images in their true colors for better accuracy. Digital cameras that act as eyes in vision tasks are deficient in, Color Constancy, an inherent property of human visual system due to which the perceived color of the objects remains constant even under varying illumination conditions. Therefore it is implicit to state that, to achieve color constancy in digital image inputs, illumination estimation is an important problem to be addressed.

\section{Color Constancy and Related Work}

\subsection{Color Constancy problem formulation}

A sample digital image in RGB color space can be simply modeled as the product of the pixels in their natural colors and the illumination present in it, as shown below in eq. \eqref{equ:Image_formulation}
\begin{eqnarray}
  I_{rgb} = W_{rgb}~~  \times ~~ L_{rgb}~
 \label{equ:Image_formulation}
\end{eqnarray}
where ~$I_{rgb}$ is the (r, g, b) tuple corresponding to each pixel, ~$W_{rgb}$ is the white balanced or 
the true color (r, g, b) tuple of each pixel and ~$L_{rgb}$ is the global illumination common across all the pixels in the image.

A true colored or white balanced image can therefore be reproduced by, first estimating the unwanted illumination present
in an image and then discounting it. Once the illumination is known, the white balanced image can be derived  from eq. \eqref{equ:Image_formulation}, as shown below in eq. \eqref{equ:White_balanced_image}
\begin{eqnarray}
  W_{rgb} = I_{rgb}~ /~ L_{rgb}~ 
  \label{equ:White_balanced_image}
\end{eqnarray}

Hence the efficiency of any Color Constancy algorithm is a measure of, how accurately it can estimate the illumination in a given image. 
The most commonly used error metric to measure the efficiency of a color constancy algorithm is ``angular error", which is defined in 
eq. \eqref{equ:Angular_Error}
\begin{eqnarray}
 \theta = cos^{-1}\left ( \dfrac{<gt, et>}{\parallel gt\parallel_{2} \cdot \parallel et \parallel_{2}}  \right )
  \label{equ:Angular_Error}
\end{eqnarray}
where ~$gt$ and ~$et$ are the ground truth illumination and the estimated illumination by an algorithm respectively.

In the following subsections, we categorize and discuss existing illumination estimation methods and differentiate the proposed method against them.

\subsection{Statistical methods}
Most of the statistical methods\cite{gonzalez2008histogram,tai2012automatic,gollanapalli2017auto} that model the color constancy task assume some regularity among the pixel colors or intensities under natural lighting conditions. For example, the grey world approach\cite{huo2006robust} assumes that the average surface reflectance in an image is gray and hence the color of illumination is the deviation from gray. Van De Weijer et al. in \cite{van2007edge} summarizes most of the statistical methods in one equation. The statistical-based methods work well on data which satisfies the prior assumptions. Their performance deteriorates exponentially when those assumptions tend to fail, resulting in poor accuracy.

\subsection{Learning based methods}
Barron in \cite{barron2015convolutional} introduced a discriminative learning based method wherein the problem of illumination estimation is reduced to a simple 2D spatial localization task.
\\
\textbf{Deep learning based methods:}
CNN based model proposed by Binaco et al. in \cite{bianco2015color} showed that CNNs are capable of capturing the distribution of how a natural image looks like, more importantly, the global illumination present in it. But this model is proven to be very compute intensive. An advanced version of CNN based model, DS-NET proposed by Shi et al. in\cite{shi2016deep} makes use of two interactive sub-networks to solve the Color Constancy problem in a better way. Its first sub-net (Hyp-Net) generates two hypothesis for illumination estimation and the second sub-net (Sel-Net) adapts to select one among them. This design increased the accuracy but resulted in a heavier model with many parameters. Hu et al. \cite{hu2017fc4} solved the problem by capturing the semantic details from the input image and introduced a novel pooling method called weighted pooling to achieve a better estimate of global illumination. It masks the estimated illuminations with the learned weight map. The weight map is learned in relative to the confidence of respective image portions contributing to the illumination. It however re-uses existing models like AlexNet \cite{krizhevsky2012imagenet} and SqueezeNet \cite{iandola2016squeezenet} that are well proven for classification tasks and hence is heavier.

In contrast, our work formulates Color Constancy task as two independent learning sub-tasks, in a multi-class learning context, that helped to achieve state-of-the-art accuracy with \textbf{30x} lesser model parameters and \textbf{70x} faster inference time.
\\

The main contributions of this paper are as follows:
\begin{enumerate}
\item We propose an efficient multi-branch architecture that independently learns the spatial contextual relationship of objects using depth-wise convolutions in one branch and color correlation among the pixels using Point-Wise convolutions in another branch. The output signals from these two branches when combined to achieve a common objective, results in better accuracy and faster inference.
\item We introduced an implicit regularization strategy based on ‘soft parameter sharing’ between the two branches of the proposed multi-branch architecture to improve upon generalization accuracy.
\item We demonstrate the applicability of proposed architecture on a class of computer vision problems such as color constancy and low-light photography.
\end{enumerate}

The claims are defended with a detailed set of visual and empirical experimental results in the results section. 

\section{Proposed Method}

\begin{figure}[h]
	\setlength{\textfloatsep}{0.1cm}
	\centering
	\includegraphics[height=2.9cm]{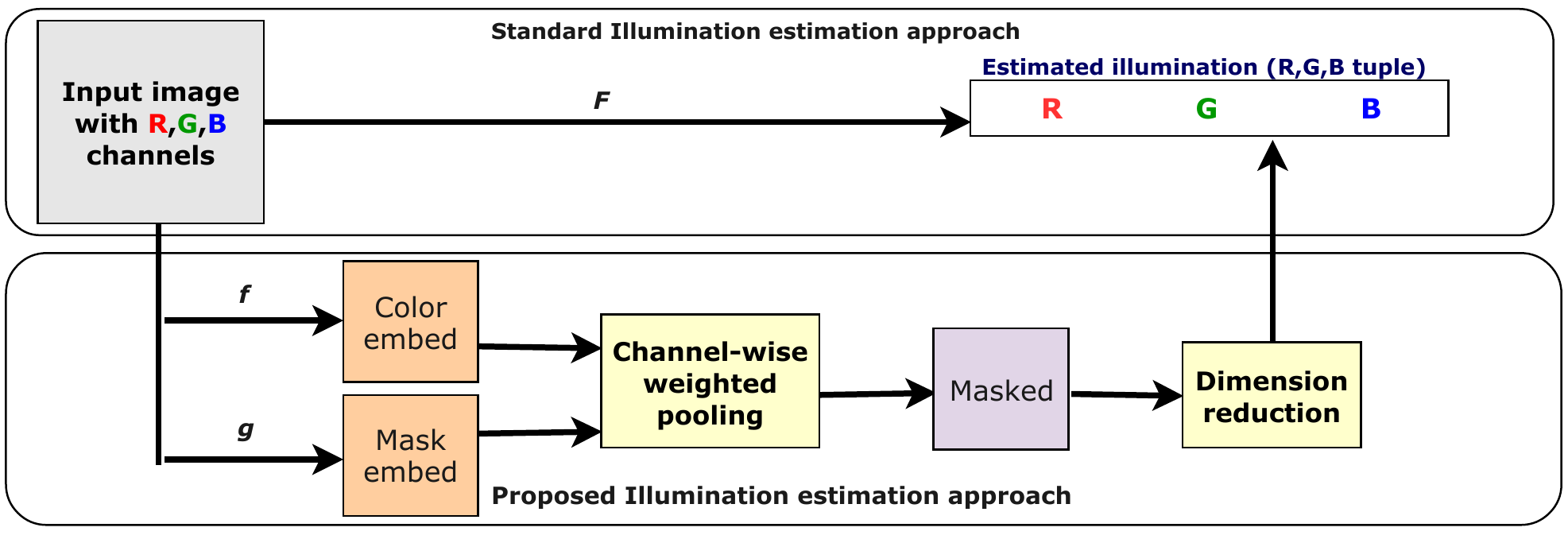}
	\caption{Overview of the proposed method}
	\label{fig:overview}
\end{figure}

This section details the proposed baseline multi-branch architecture. The problem of illumination estimation has been modeled as extracting color properties present at the low-level image pixels and masking those with contextually rich image regions that can provide useful semantic information. The masked color representation is further transformed to match the required output dimension. An overview of our approach is shown in the fig. \ref{fig:overview}. We also present the implicit regularization technique used to retain the high generalization accuracy for all our use-cases.
\subsection{Extracting Color property}
We denote the color property extraction operation by function $ f:\mathbb{R}^{(H\times W\times 3)}\rightarrow \mathbb{R}^{(h\times w\times k)} $ where $h<H, w<W$ and $k>=3$. $f$ transforms the input image tensor from RGB space to an intermediate embedding space. It outputs a reduced spatial dimension tensor that has per-pixel color correlation in the input image. In order to approximate $f$, we use Point-wise Convolution operator. The notation of $f$ is formalized below.
\begin{equation}
  PointwiseConv(W,y)_{i,j} = \sum_{m}^{M}W_{m}\cdot y_{i*s,j*s,m}
  \label{equ:pointwise}
\end{equation}
where $y$ is a input tensor having spatial dimension of $(K,L)$ and depth dimension of $M$. $s$ is the stride used in the convolution. A choice of $s = 2$ reduces the spatial dimension to $(K/2, L/2)$. $W$ is the learn-able weights with dimension $(1,1,M)$.

We apply $2\times M$ point-wise convolution with learn-able weights $\left \{W^{1}, W^{2}, ..., W^{2\times M}\right \}$ that produces an output tensor with dimension $(K/2, L/2, 2\times M)$. A unit block that extracts color properties present in the input tensor is then given by eq. \eqref{equ:unit_color}
\begin{equation}
f_{c}(y) = \sigma (PointwiseConv(W^{1:2M},y))
  \label{equ:unit_color}
\end{equation}
where $\sigma$ is a non-linearizing unit $RELU$.
Using eq. \eqref{equ:unit_color}, $f$ is defined in eq. \eqref{equ:color}
\begin{equation}
f(I) = f_{c}^{N}(I)
  \label{equ:color}
\end{equation}
where $I$ is the input image and $N$ is the number of repeated units of color property extraction blocks.

\subsection{Extracting Semantic Map}
We define the function $g:\mathbb{R}^{(H\times W\times 3)}\rightarrow \mathbb{R}^{(h\times w\times k)}$ that generates the contextual mask based on the input image. It can be observed that to extract the  contextual information, depth dimension is not required. Instead, we focus only along the spatial dimension and hence utilize Depth-wise convolution operator for extracting the spatial contextual information. The notion is formalized below. 
\begin{equation}
  DepthwiseConv(W,y)_{i,j,k} = \sum_{p,q}^{P,Q}W_{p,q}^{k''}\cdot  y_{i+p,j+q,k'}
  \label{equ:depthwise}
\end{equation}
where $y$ is the input tensor and $W$ is the learn-able weight with dimension $(P,Q)$. $k$ varies from $1$ to $2*K$. $k' = \left \lfloor k/2 \right \rfloor$. $k'' = k_{mod(k,2)}$ or in other words, we have set a depth-multiplier of depth-wise convolution as 2. $h_{s}$ denotes a unit block that extracts semantic information from the input tensor, define in eq. \eqref{equ:unit_map2}
\begin{equation}
h_{s}(y) = \lambda (\sigma (DepthwiseConv(W^{1:2M},y)))
  \label{equ:unit_map2}
\end{equation}
where $\lambda$ is pooling unit $AVGPOOL$. The output dimension of the embedding is $(K/2, L/2, 2*M)$.
The semantic map generation function is then defined by eq. \eqref{equ:map2}
\begin{equation}
g(I) = h_{s}^{N}(I)
  \label{equ:map2}
\end{equation}

\subsection{Channel-wise weighted pooling}
In order to apply the semantic map on the color embedding, we propose a novel pooling technique called channel-wise weighted pooling. Like weighted pooling in \cite{hu2017fc4}, the proposed channel-wise weighted pooling does not constraint the number of weighting masks to one, instead it provides a mask for each channel. Also the number of color channels can be more than three, which is later reduced to match the required output channel dimension. Intuitively, this gives each learned color properties more flexibility to select image regions before aggregating for a global property. The masked output $O$ is the element-wise product of the signals from $f$ and $g$ respectively and is shown in eq. \eqref{equ:pool}.

\begin{equation}
  O^{k}_{ij} = S_{ij}^{k} * C_{ij}^{k}
  \label{equ:pool}
\end{equation}

$S = f(I)$ and $C = g(I)$ are the respective output embedding tensors with spatial dimension along $(i,j)$ and depth dimension along $k$.

The output embedding $O$ from eq. \eqref{equ:pool} is transformed to the required output dimension by spatial reduction followed by depth reduction. The reduced output is then normalized to obtain illumination estimation as shown in eq. \eqref{equ:reduce2}
%\begin{equation}
%  O^{k} = \sum_{i=1}^{N}\sum_{j=1}^{M} O_{ij}^{k}
%  \label{equ:reduction}
%\end{equation}
\begin{eqnarray}
&O^{k} = \sum_{i=1}^{N}\sum_{j=1}^{M} O_{ij}^{k}
\\
&O^{i}_{reduced} = \sum_{k=i*K/3}^{(i+1)*K/3} O^{k}, i = 0,1,2
\\
&(I_{r}, I_{g}, I_{b}) = normalization(O_{reduced})
   \label{equ:reduce2}
\end{eqnarray}

The proposed baseline architecture is shown in fig. \ref{fig:arch_branch} below. The architecture is end-to-end trainable and does not require supervisory signals for $S$ and $C$ independently. The output $(I_{r}, I_{g}, I_{b})$ is optimized with respect to ground truth. With the above formulation, each of the branches learns a specific function, namely semantic feature in branch-1 and color correlation information in branch-2.

\begin{figure}[h]
	\centering
	\includegraphics[height=4.5cm]{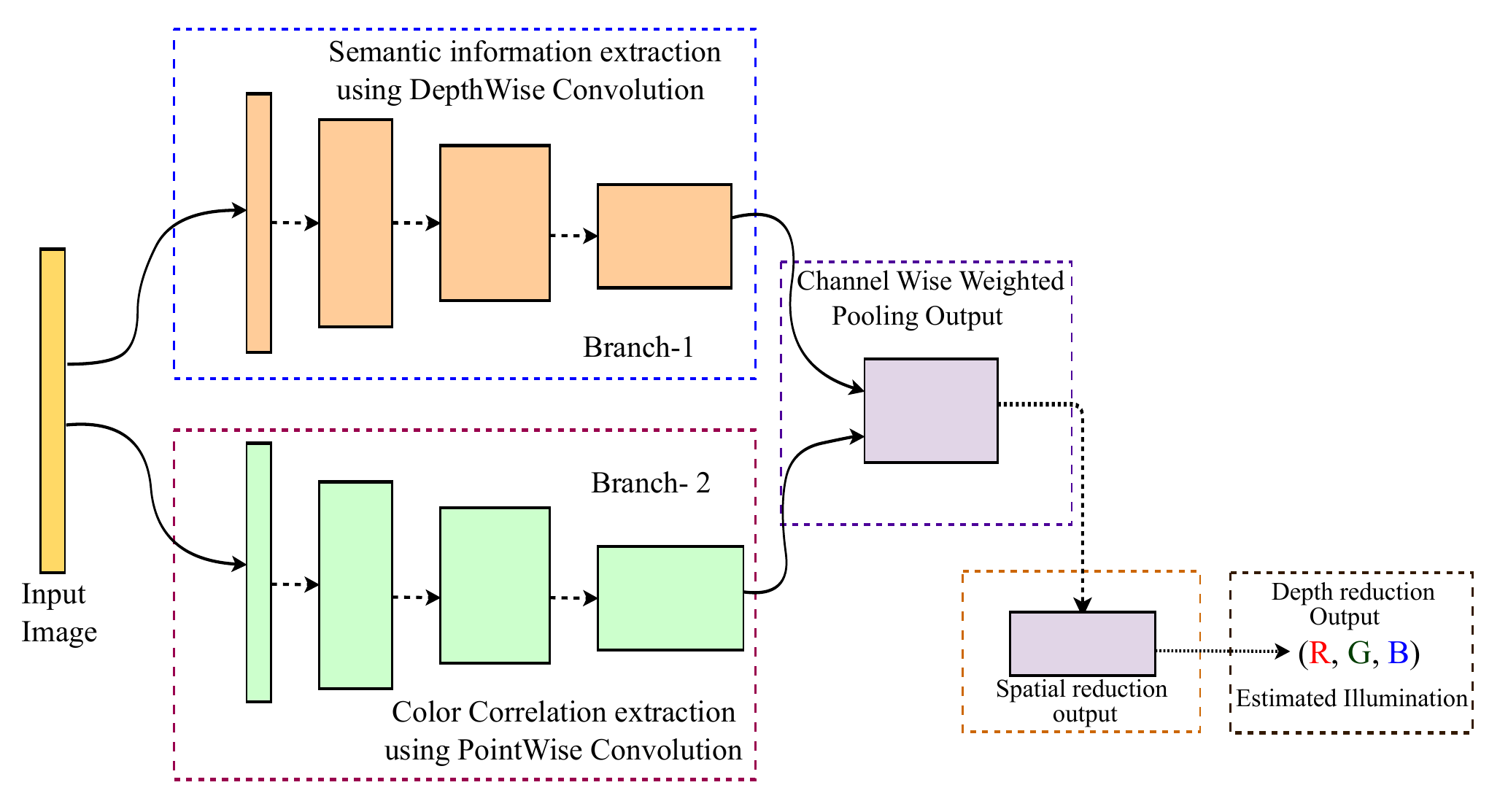}
	\caption{Baseline dual branch architecture overview. Input image is convolved with depth-wise convolution along branch 1 and point-wise convolution along branch 2. The shape of intermediate tensors are kept same in both branches to retain Fully Convolution Property(FCN). Spatial dimension is halved and depth dimension is doubled at every layer. Input image is convolved with a 3X3 convolution with 32 filters before passing the input along task specific branches}
	\label{fig:arch_branch}
\end{figure}

\subsection{Implicit Regularization with soft parameter sharing}
In the formulation above, $g$ in eq. \eqref{equ:map2} can be thought of as an auxiliary task in a multi-class learning problem \cite{argyriou2007multi} that aids in selecting best regions for each color representation extracted by $f$ as shown in eq. \eqref{equ:color}. Intuitively, the weighted mapping helps in focusing on regions that have rich contextual information while obscuring irrelevant regions before estimating for a global property. However, the two signals produced from $f$ and $g$ are not completely independent, they are loosely dependent, as they are required to learn for a common objective jointly. Having these two tasks learn their specific signals independently pose overfitting problems \cite{hawkins2004problem} since available datasets are not large enough.

In general, to combat overfitting problem, regularization techniques \cite{kukavcka2017regularization} such as $L1$ regularization or $L2$ regularization, among others can be used. These are explicit regularization techniques that adds a regularizer term in the optimization function 

\begin{figure}[h]
	\centering
	\includegraphics[height=3.9cm]{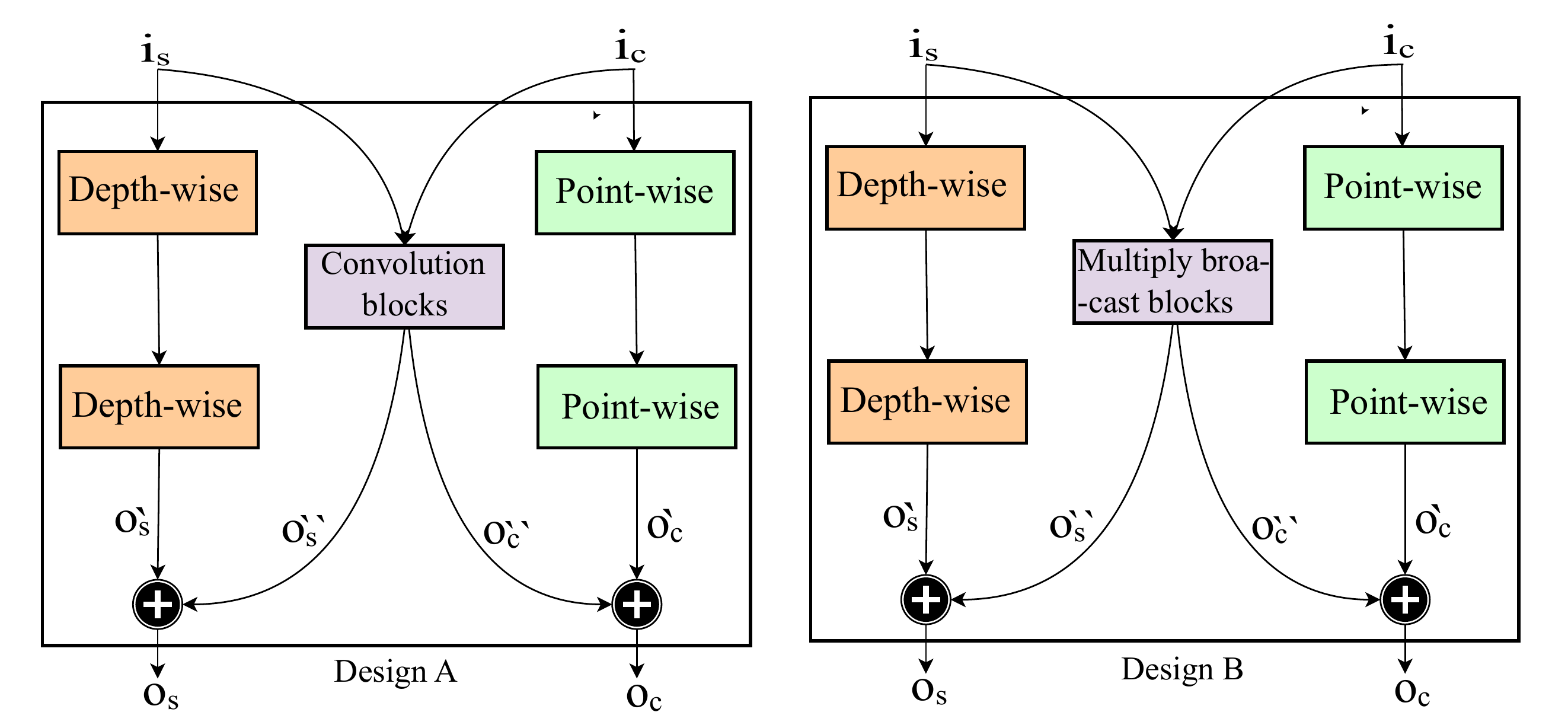}
	\caption{Proposed two designs to achieve regularization. Parameters between two task specific branches are shared as shown in designs A and B. Design A has three convolution operators, among which two are  input specific and one is shared across both inputs. Similarly, Design B uses two weight scalars corresponding to two inputs and one shared scaler to combine the input signals. Unlike convolution unit, the scaler weights impose stricter constraints. Inputs of shape $(H,W,D)$ is transformed to $(H/4, W/4, 2*D)$ by each such regularizing units. For clarity, non-linearization and pooling are not shown.}
	\label{fig:Blocks}
\end{figure}

In contrast, we present a novel micro-architecture design for regularizing baseline multi-branch architecture. The proposed design blocks are shown in fig. \ref{fig:Blocks}. Sharing parameters between two signals extractor $f_{S}$ and $f_{C}$ is a result of our observation that imposing assumptions of task dependencies in the architecture provide some inductive bias.

In other words, these architectural changes work because our re-constructed hypothesis space $\mathcal{H}$ puts constraints on the estimator $f$ and $g$. The constraint being soft parameter sharing between two branches, and thus effectively shrinks the class $\mathcal{H}$. The network is trained in an end-to-end manner and thus optimizer searches through $\mathcal{H}$ automatically for a good estimator $f$ and $g$, or identifies how much to share at which layer. In eq. \eqref{equ:sharing} through \eqref{equ:main2}, we formalize the regularization with soft parameter sharing.
\begin{gather}
\begin{bmatrix} o_{s}''\\  o_{c}''
\end{bmatrix}
=
\begin{bmatrix}
	w_{s} & w_{sc}\\
	w_{sc} & w_{c}
\end{bmatrix}\circledast
\begin{bmatrix}
	i_{s}\\
	i_{c}
\end{bmatrix}
\label{equ:sharing}
\end{gather}
where $i_{s}$ and $i_{c}$ are input signals of tasks specific to spatial contextual information and color representation respectively. $w_{s}$ and $w_{c}$ are learnable parameter corresponding to input signals $i_{s}$ and $i_{c}$ respectively. $w_{sc}$ is a shared parameter that combines $i_{s}$ and $i_{c}$ signals. $o_{*}''$ represents the combined output signal.
$\circledast$ denotes a convolution operator (Design A) or a scaler broadcasting product (Design B). Additions are done with element-wise addition operator.
\\
Branch specific computation are specified by \eqref{equ:out1} and \eqref{equ:out2}.
\begin{eqnarray}\label{equ:out1}
  o_{s}' = h_{s}^{2}(i_{s})
\\
  \label{equ:out2}
  o_{c}' = f_{c}^{2}(i_{c})
\end{eqnarray}
where $o_{*}'$ represents output signals corresponding to the branch specific tasks. $h_{s}$ and $f_{c}$ as defined in eqs. \eqref{equ:map2} and \eqref{equ:color} respectively.

Output of the block is then simply given by $o_{s}$ and $o_{c}$ as shown in eqs. \eqref{equ:main1} and \eqref{equ:main2}.
\begin{eqnarray}\label{equ:main1}
  o_{s} = o_{s}' \oplus o_{s}''
\\
  \label{equ:main2}
  o_{c} = o_{c}' \oplus o_{c}''
\end{eqnarray}
where $\oplus$ is a concatenation operator along depth dimension.
We discuss different design strategies below:
\begin{itemize}
\item{\textbf{Parameter sharing with a convolution operator (Design A)}}: Convolution operators linearly transform the two signals. If there is no sharing of parameters between two input signals, then $w_{sc}$ will be a zero tensor. With this design, we achieve state-of-the-art accuracy on color constancy task with far lesser computation time.

\item{\textbf{Parameter sharing with scalar weights (Design B)}}: The scalers scales the input signals. If there is no sharing of parameters, then $w_{sc}$ will be a zero scaler. This design achieves an accuracy which is under visual acceptable limits, but is the lightest and fastest.
\end{itemize}
In the two designs A and B, we observe two architectural characteristics:
\begin{enumerate}
\item Soft parameter sharing that helps the architecture in reducing generalization error.
\item Merging of task specific signals helps in easy flow of gradients. They mimic the behavior of residual connections \cite{he2016identity} \cite{he2016deep}.
\end{enumerate}
As shown by He. et. al. \cite{he2016deep}, a residual connection breaks the output $H(x)$ to $F(x)+x$ or $F(x)+Wx$. Under this settings, eqs. \eqref{equ:identity} and \eqref{equ:trans_identity} formalizes the outputs from both designs.
\begin{eqnarray}\label{equ:identity}
  out = \mathcal F(in) + in
\\
  \label{equ:trans_identity}
  out = \mathcal F(in) + W\cdot in
\end{eqnarray}
where $\mathcal F(in) = DepthwiseConv(W,h_{s}(in))$ if input is spatial signal, else $\mathcal F(in) = PointwiseConv(W,f_{c}(in))$.
\\

Eq. \eqref{equ:identity} and \eqref{equ:trans_identity} are two special cases which may arise in eq.  \eqref{equ:sharing}. We observe that when when $w_{sc}$ is a zero kernel (Design A) or a zero weight scaler (Design B), the case is covered by the eq. \eqref{equ:trans_identity}. Over the previous condition, when $w_{s}$ and $w_{c}$ are identity kernel (Design A) or a unit scaler (Design B), then this case is shown by eq. \eqref{equ:identity}.

Next section details the improvement results on color constancy and low-light photography with the above discussed design choices.

%--------------------------------------------------------------------------------------------------------------
\section{Experiments}

\subsection{Experiments and evaluation on color constancy}

This section evaluates the proposed method in terms of efficiency and accuracy for color constancy task on Cube\cite{Banic2018UnsupervisedLF} and 
NUS-8 \cite{cheng2014illuminant} data sets. Our baseline architecture and its regularized versions as shown in Fig. \ref{fig:arch_branch} and Fig. \ref{fig:Blocks} are implemented in tensorflow framework~\cite{abadi2016tensorflow}. 
The details of the architecture are described in the Table. \ref{table:Arch_overview}.

%------------------------------Architecture table-----
\setlength{\tabcolsep}{2pt}
\setlength{\textfloatsep}{0.1cm}
\begin{table}[h]
\fontsize{7}{10}\selectfont
\begin{center}
\caption{
Table describes architecture details for color constancy problem. We use Design A or B as regularization blocks. The baseline multi-branch architecture can be constructed from either of the designs by removing the parameter sharing block.
}
\label{table:Arch_overview}
\begin{tabular}{llllll}
\hline\noalign{\smallskip}
Input-Output Tensor & Tensor shape & Operation & Strides & Filters &\\
\noalign{\smallskip}
\hline
\noalign{\smallskip}
Input  & (512,512,3) & conv 3X3 & 2 & 32 &\\
Tensor1  & (256,256,32) & Design A or B & --- & --- &\\
Tensor2a,Tensor2b  & (64,64,64) & Design A or B & --- & --- &\\
Tensor3a,Tensor3b  & (16,16,128) & Channel-wise weighting & --- & --- &\\
Tensor4  & (16,16,128) & Spatial reduction & --- & --- &\\
Tensor5  & (1,1,128) & Channel depth reduction & --- & --- &\\
Output tensor & (1,1,3) & --- & --- & ---&\\
\hline
\end{tabular}
\end{center}
\end{table}
\setlength{\tabcolsep}{1.0pt}

The training was done in end-to-end manner on a workstation with Nvidia GeForce GTX-1080Ti GPUs. 
We use Adam optimizer with a batch size of 32, and a learning rate of $5*10^{-4}$ for all of the designs. 
No explicit regularization techniques such as $L1$ or $L2$ has been used. For optimization, we use Mean Squared Error (MSE) of normalized ground truth illumination and the estimated illumination predicted by the network, 
while for performance evaluation, we use angular error as given by eq. \eqref{equ:Angular_Error}.

We augment data on both Cube and NUS-8 datasets, by randomly cropping and flipping the images along horizontal and vertical axis.  
All images in test and training sets are gamma corrected  with a $\gamma$ of $1/2.2$. The Cube portion and color checker from Cube and NUS-8 datasets respectively are masked, for both training and testing.

%----------------------------Table 1--------------------------------------------------

\setlength{\tabcolsep}{4pt}
\setlength{\textfloatsep}{0.1cm}
\begin{table}[h]
\fontsize{7}{10}\selectfont
\begin{center}
\caption{
Results on NUS-8 data set. Number of parameters are given in Millions. Floating point operations are given in Giga-Flops. DS-Net* is evaluated with an input shape of $(47, 47, 2)$ while others are evaluated with an input shape of $(224, 224, 3)$.
}
\label{table:NUS}
\begin{tabular}{lllllllll}
\hline\noalign{\smallskip}
Models & Mean & Median & \multicolumn{1}{p{0.5cm}}{\centering Tri \\ mean} & \multicolumn{1}{p{0.5cm}}{\centering Best \\ 25\%} & \multicolumn{1}{p{0.5cm}}{\centering Worst \\ 25\%} & Params & Flops\\
\noalign{\smallskip}
\hline
\noalign{\smallskip}
Grey-world  & 4.14 & 3.2 & 3.39 & 0.9 & 9 & -- & -- &\\
DS-Net  & 2.24 & 1.46 & 1.68 & 0.48 & 5.28 & 2.64 & 0.031* &\\
FC4-alex  & 2.12 & 1.53 & 1.67 & 0.48 & 4.78 & 2.9 & 1.2 &\\
FC4-squeeze  & 2.23 & 1.57 & 1.72 & 0.47 & 5.15 & 1.23 & 0.77 &\\
\textbf{Design A}  & \textbf{2.102} & 1.654 & \textbf{1.72} & 0.576 & \textbf{4.469} & \textbf{0.13} & \textbf{0.11} &\\
\textbf{Design B}  & 2.442 & 1.871 & 1.956 & 0.67 & 5.283 & \textbf{0.04} & \textbf{0.04} &\\
\hline
\end{tabular}
\end{center}
\end{table}
\setlength{\textfloatsep}{0.1cm}
With three-fold cross validation, we compare our results with other methods on standard metrics such as mean, 
median, tri-mean, mean of the lowest 25\%, and mean of highest 25\%. Results with NUS-8 dataset is presented in Table \ref{table:NUS}. 
It is to be noted that, as compared to state of the art learning based methods such as FC4~\cite{hu2017fc4}, 
accuracy is comparable and the number of computation cycles are reduced by ~70 times. 
%----------------------------------------Table 2-------------------------------------------
\setlength{\tabcolsep}{4pt}
\setlength{\textfloatsep}{0.05cm}
\begin{table}[h]
\fontsize{7}{10}\selectfont
\begin{center}
\caption{
Comparison study of Color Constancy on Cube data set. As compared to baseline multi-branch architecture, regularization using Design A improves the accuracy significantly. More visual results are provided in the supplement.
}
\label{table:Cube}
\begin{tabular}{lllllll}
\hline\noalign{\smallskip}
Models & Mean & Median & Tri-mean & Best 25\% & Worst 25\%\\
\noalign{\smallskip}
\hline
\noalign{\smallskip}
Grey-world  & 3.75 & 2.91 & 3.15 & 0.69 & 8.18 &\\
Color Tiger  & 2.94 & 2.59 & 2.66 & 0.61 & 5.88 &\\
Restricted Color Tiger & 1.64 & 0.82 & 1.05 & 0.24 & 4.37 &\\
\textbf{Baseline}  & 1.701 & 1.111 & 1.276 & 0.345 & 4.003 &\\
\textbf{Design A} & \textbf{1.616} & 1.09 & 1.242 & \textbf{0.318} & \textbf{3.76} &\\
\hline
\end{tabular}
\end{center}
\end{table}

\begin{figure}[H]
	\centering
	\includegraphics[height=8.0cm]{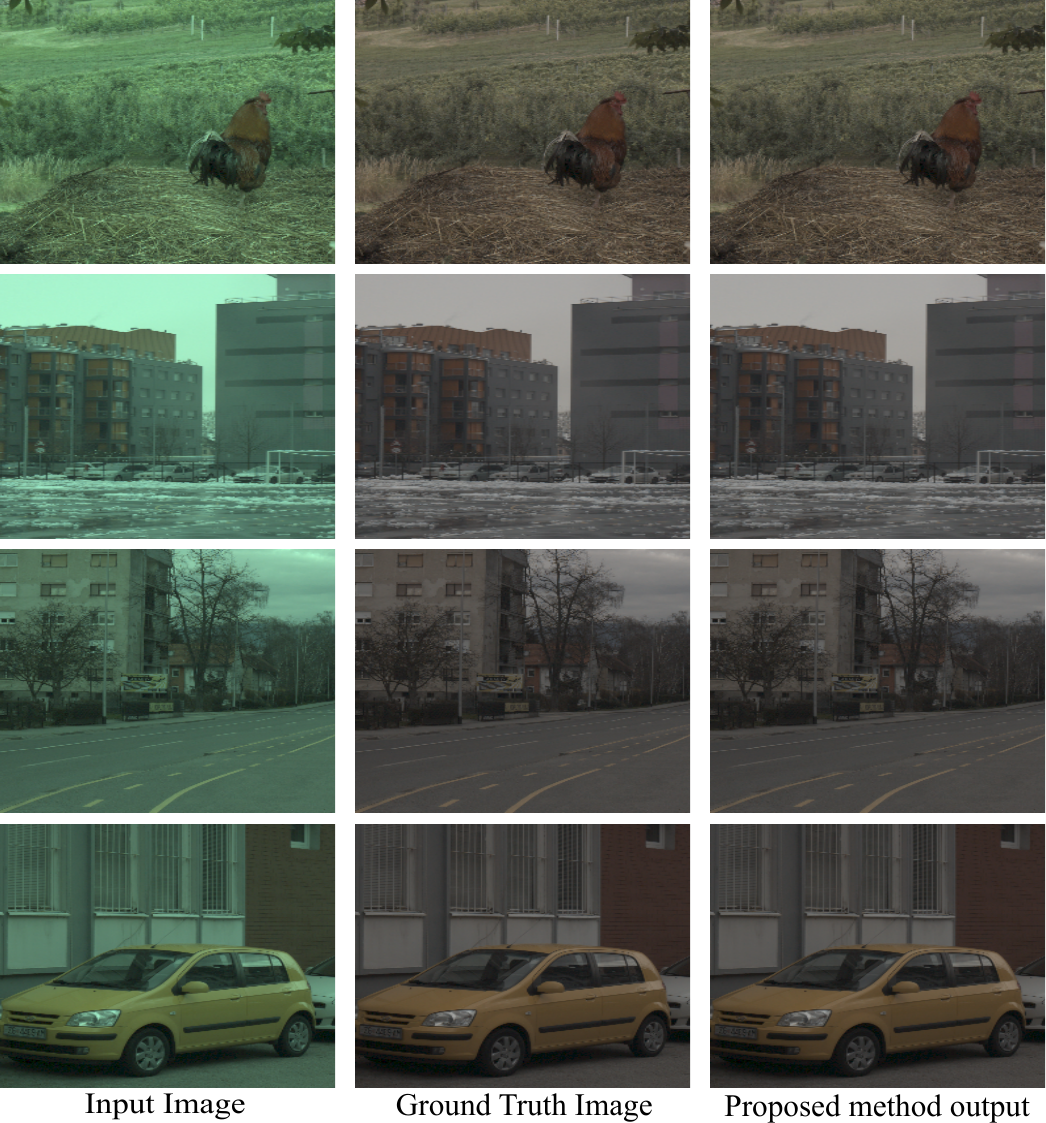}
	\caption{Sample visual results on Cube dataset with respective angular error of  0.115, 0.081, 0.259 and 0.167 degrees.}
	\label{fig:CC_results}
\end{figure}

Effect of our designs in regularizing the baseline model is being reflected in the results on as shown in Table \ref{table:Cube}. Better results on mean of worst 25\% metric (hard to learn examples) in both the tables shows the robustness of our model due to regularization method.
The difference in accuracy between baseline and regularized versions defends our analysis. For even better efficiency, quantization and sparsity reduction methods can be used. The inference time of our model on a single threaded ARM based platform running at 2.1 GHz is \textbf{30ms} as compared to \textbf{100ms}
for FC4-SqueezeNet model. Hence, the proposed model can be very well used to realize real time mobile based applications.

%----------------------------------------------------------------------------
\subsection{Feasibility demo for low-light photography}

This section demonstrates, how the proposed architecture can be a simple plug and play for low-light photography use-case and compares the results against Learning to See in the Dark \cite{Chen_2018_CVPR}, which is the state-of-the-art DNN based method.

\begin{figure}[H]
	\centering
	\includegraphics[height=4.7cm]{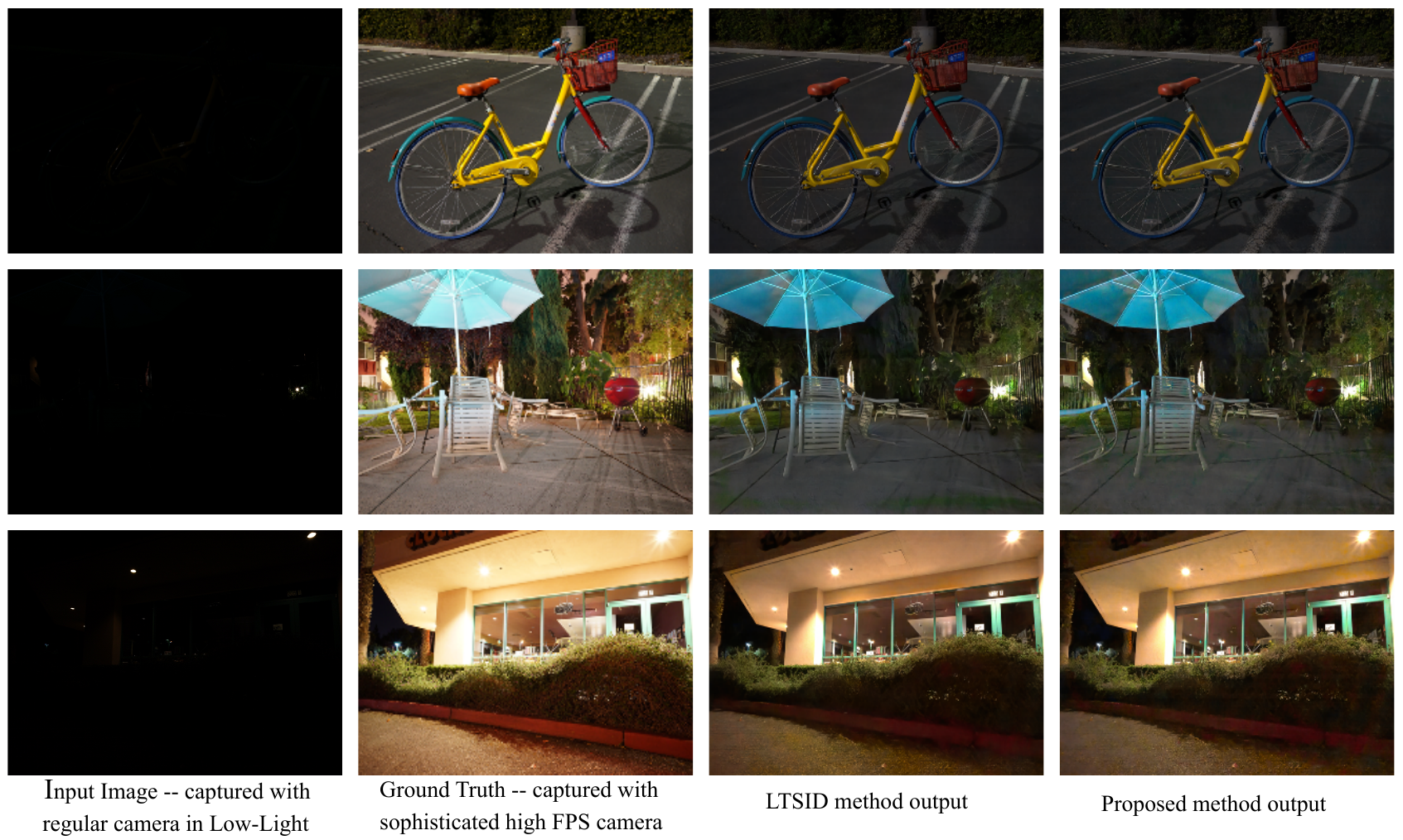}
	\caption{Sample visual results for SID dataset}
	\label{fig:low_light}
\end{figure}

Fig. \ref{fig:low_light} depicts sample visual comparison results of the Learning to See in the Dark (LTSID) and proposed method on Sony camera images in See-in-the-Dark (SID) dataset.

By simply replacing the convolution blocks in the encoder part of the U-Net architecture in LTSID with our proposed design A, we achieve the above shown visual accuracy with ~50\% lesser parameters.

\section{Conclusion}

In this paper, we propose to solve a class of computer vision problem such as color constancy. We presented an efficient multi-branch architecture that utilizes depth-wise convolution to learn semantic features and point-wise convolution to learn color correlation. With channel-wise weighted pooling layer, we combine the two
signals in order to predict a global property. To improve the accuracy of our method, we introduce an implicit regularization technique based on multi-task soft parameter sharing. The experimental results confirm that with the design choices as ours, the method can lead to a higher accuracy in tasks such as illumination estimation while being under computational limits.

% References should be produced using the bibtex program from suitable
% BiBTeX files (here: strings, refs, manuals). The IEEEbib.bst bibliography
% style file from IEEE produces unsorted bibliography list.
% -------------------------------------------------------------------------
\bibliographystyle{IEEEbib}
\bibliography{decouplesematic}

\end{document}